\documentclass[10pt,twocolumn,letterpaper]{article}

\usepackage{iccv}              % To produce the CAMERA-READY version
\usepackage{float}
\definecolor{iccvblue}{rgb}{0.21,0.49,0.74}
\usepackage[pagebackref,breaklinks,colorlinks,allcolors=iccvblue]{hyperref}
\usepackage{algorithm}
\usepackage{algorithmic}
\def\paperID{2} % *** Enter the Paper ID here
\def\confName{ICCV}
\def\confYear{2025}

\title{Best Foot Forward: Robust Foot Reconstruction in-the-wild}

\author{Kyle Fogarty\\
University of Cambridge\\
{\tt\small ktf25@cam.ac.uk}
\and
Jing Yang\\
University of Cambridge\\
{\tt\small jy496@cam.ac.uk}
\and
Chayan Kumar Patodi\\
Hike Medical\\
{\tt\small chayan@hikemedical.com}
\and
Jack Foster\\
Hike Medical\\
{\tt\small jack.foster@hikemedical.com}
\and
Aadi Bhanti\\
Hike Medical\\
{\tt\small aadi@hikemedical.com}
\and 
Steven Chacko\\
Hike Medical\\
{\tt\small steven@hikemedical.com}
\and
Cengiz Öztireli\\
University of Cambridge\\
{\tt\small aco41@cam.ac.uk}
\and
Ujwal Bonde\\
Hike Medical\\
{\tt\small ujwal@hikemedical.com}
}

\begin{document}

\maketitle
% \begin{abstract}
% Accurate 3D foot reconstruction is crucial for personalized orthotics, digital healthcare, and virtual fittings. However, existing methods struggle with incomplete scans and anatomical variations, particularly in self-scanning scenarios where user mobility is limited, making it difficult to capture areas like the arch and heel. We present a novel end-to-end pipeline that refines Structure-from-Motion (SfM) reconstruction. It first resolves scan alignment ambiguities using SE(3) canonicalization with a viewpoint prediction module, then completes missing geometry through an attention-based network trained on synthetically augmented point clouds. Our approach achieves state-of-the-art performance on reconstruction metrics while preserving clinically validated anatomical fidelity. By combining synthetic training data with learned geometric priors, we enable robust foot reconstruction under real-world capture conditions, unlocking new opportunities for mobile-based 3D scanning in healthcare and retail. We will release our dataset online at: \href{https://bestfootforward.netlify.app}{\texttt{https://bestfootforward.netlify.app}}.
% \end{abstract}

\begin{abstract}
High-fidelity 3D foot reconstruction is crucial for prescription orthotics but is hindered by expensive, specialized equipment that limits patient access. We overcome this barrier with the first end-to-end pipeline to reconstruct clinically-accurate foot meshes from simple, self-captured smartphone videos. Our method uniquely solves the core challenges of in-the-wild scanning: we resolve pose ambiguities using SE(3) canonicalization with viewpoint prediction, and then complete partial geometry using an attention-based network. Clinical validation demonstrates that our reconstructions achieve state-of-the-art accuracy and meet prescription-readiness standards, preserving the anatomical fidelity essential for medical intervention. By democratizing high-quality foot assessment, our work unlocks new opportunities for accessible telemedicine, preventative diabetic care, and personalized orthotic treatment.
We will release our dataset online at: \href{https://bestfootforward.netlify.app}{\texttt{https://bestfootforward.netlify.app}}.
% High-fidelity 3D foot reconstruction is essential for prescription orthotics where accuracy directly impacts patient outcomes. 
% Existing clinical methods depend on expensive equipment, controlled environments, and skilled operators, significantly limiting access to quality foot assessment. 
% We present the first end-to-end pipeline to reconstruct clinically accurate full-foot meshes from unstructured self-capture videos. 
% Our approach democratizes foot assessment by resolving the fundamental challenges of self-scanning: we address scan alignment ambiguities using SE(3) canonicalization with viewpoint prediction, then complete missing geometry through an attention-based network trained on synthetically augmented point clouds. 
% Clinical validation shows our reconstructions achieve state-of-the-art accuracy and meet prescription-readiness standards, while preserving the anatomical fidelity crucial for intervention.
% By enabling robust foot reconstruction under real-world capture conditions, our method unlocks new opportunities for telemedicine, preventative diabetic care, and accessible orthotic treatment. 
% We will release our dataset online at: \href{https://bestfootforward.netlify.app}{\texttt{https://bestfootforward.netlify.app}}.
\end{abstract}

\begin{figure*}
    \centering
     \includegraphics[width=\linewidth]{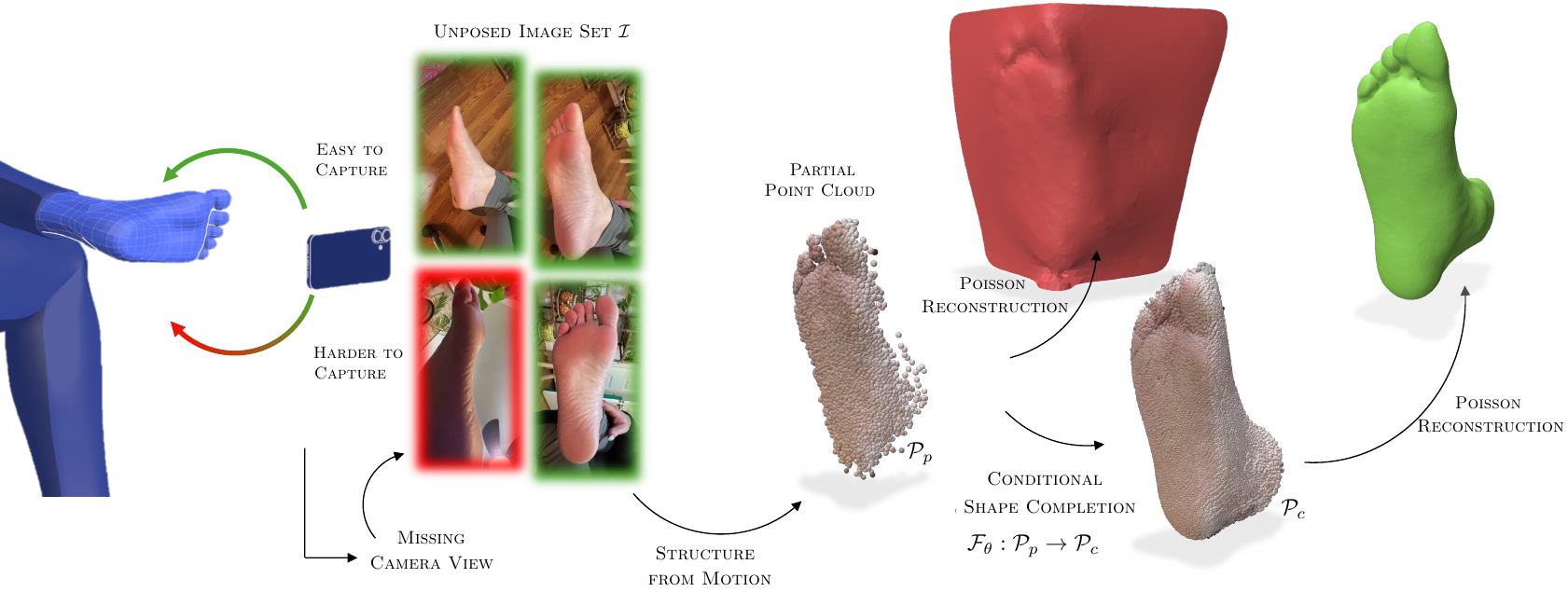}
     \caption{Challenges in foot self-scanning for individuals with reduced mobility: The image highlights the difficulty of capturing the complete foot geometry, especially the underside (red regions), which is harder to access; this limitation often leads to incomplete foot geometry.}
    \label{fig:Self-Scan}
\end{figure*}    
\section{Introduction}

High-fidelity digital twins of the human body represent a foundational technology in autonomous healthcare, with the capacity to deliver significant advancements in personalized medicine. Toward this goal, we present a novel method for high-quality foot geometry capture, tailored to the design of custom foot orthotics. The proposed vision-based workflow automates the acquisition of complex foot geometries, demonstrating robustness against incomplete capture, to enable a fully digital manufacturing process consistent with the tenets of personalized medicine \cite{leite2019design}. The methodology has direct implications for a spectrum of applications, including custom prosthetics, preoperative planning, and immersive digital environments such as virtual try-ons and gaming.\\

{
While significant progress has been made in reconstructing human anatomy from images \cite{bogo2016keep, loper2023smpl, kanazawa2018end, keller2023skin}, creating accurate models of deformable and self-occluding parts like the human foot remains a formidable challenge. Self-captured scanning with commodity devices remains an unsolved problem while traditional approaches, including laser scanners \cite{silva2024review}, structured-light booths \cite{gaertner1999multiple}, marker-based gait laboratories \cite{moro2022markerless}, and impression boxes \cite{leung2004orthotic}, rely on costly hardware, controlled environments, and skilled operators. % TODO: Add citations to this sentence.
Although recent efforts have sought to democratize high-quality foot reconstruction via smartphone video {\cite{kok2020footnet, boyne2024found}}, these approaches often depend on LIDAR-equipped devices, limiting accessibility, or they are not robust to in-the-wild partial scans, leading to unreliable geometry. To address this gap, we present the first method that \textit{robustly} reconstructs high-quality foot geometry from a single, self-captured video taken with a standard smartphone.\\}

% Legacy solutions, including laser scanners, structured-light booths, marker-based gait labs, and impression boxes, depend on expensive hardware, controlled settings, and expert operators. 

% . This explains their limited real-world adoption despite decades of research. To our knowledge, we are the first to reconstruct a clinically accurate, full-foot mesh, including the hidden plantar arch, from unstructured, handheld smartphone video.

% Despite advances in human body reconstruction \cite{loper2023smpl,keller2023skin}, the foot remains largely unexplored due to its complex biomechanics, high morphological variance, and imaging challenges like plantar surface occlusion. To address this, we present a novel, high-quality foot reconstruction method using multi-view mobile phone images, offering a robust and accessible solution for clinical and commercial use.\\

While classical 3D reconstruction pipelines based on Structure-from-Motion (SfM) and Multi-View Stereo (MVS) excel under controlled conditions \cite{haming2010structure}, they are notoriously brittle for in-the-wild medical capture. The self-scanning of a foot is a particularly challenging instance of this problem; user-captured sequences are often sparse and incomplete due to limited mobility (Fig. \ref{fig:Self-Scan}). Without a strong geometric prior, the resulting reconstructions are not suitable for designing effective orthotics. To overcome this fundamental limitation, we introduce a learning-based framework that robustly infers complete, high-fidelity foot geometry from imperfect data. Our primary contributions are:

\begin{enumerate}
    \item {A {novel end-to-end pipeline for foot capture and completion}, capable of accurately inferring the occluded geometry from partial and noisy point clouds.}
\item The {introduction of a comprehensive dataset of 3D foot scans} (\texttt{Hike3D}), featuring broader demographic diversity than existing public datasets.
    % \item {An end-to-end reconstruction pipeline} that combines scan canonicalization, viewpoint estimation, and geometry completion to generate watertight, anatomically faithful meshes from self-captured inputs
    \item{Extensive experiments demonstrating that our method significantly outperforms both classical pipelines (COLMAP) and state-of-the-art neural rendering techniques (e.g., 3D Gaussian Splatting) in reconstruction robustness, geometric accuracy, and orthotic design suitability.}
\end{enumerate}

% Building on advances in 3D computer vision, Structure-from-Motion (SfM) enables 3D reconstruction from 2D image sequences, while Multi-View Stereo (MVS) enhances geometric detail under controlled conditions \cite{haming2010structure}. However, self-scanning the foot remains challenging—users struggle to capture dense, overlapping views due to limited mobility and awkward angles, leading to incomplete image coverage (see Fig. \ref{fig:Self-Scan}).
% To address this, we make the following key contributions:\\
% \begin{enumerate}
%     \item {The first foot completion network to refine incomplete scans, improving robustness and accuracy in foot reconstruction.}
%     \item {A diverse foot dataset \texttt{[REDACTED]} with greater variation in attributes like age and height than previous datasets, enabling more robust modeling across different foot shapes.}
%     \item {Seamless integration with template-based foot reconstruction methods to generate high-quality meshes from partial point clouds.}
%     \item {A comprehensive evaluation showing our method outperforms COLMAP and state-of-the-art Gaussian splatting in robustness, feature accuracy, and surface quality.\\}
% \end{enumerate}

\noindent
\section{Related Work}

\noindent
\textbf{Multi-view reconstruction:} Recovering 3D geometry from multiple images fundamentally depends on accurate estimation of camera intrinsics and relative poses. These parameters can be acquired through onboard sensors such as Inertial Measurement Units (IMUs) or estimated via sparse reconstruction techniques like Structure-from-Motion (SfM)~\cite{schoenberger2016sfm}. Once the camera parameters are established, dense reconstruction is typically performed using Multi-View Stereo (MVS)~\cite{schoenberger2016mvs}, which computes consistent depth and normal maps across multiple views to produce an oriented point cloud. This point cloud is then converted into a surface mesh using a reconstruction method such as Poisson Surface Reconstruction~\cite{kazhdan2006poisson}. A widely used implementation of this full pipeline is COLMAP~\cite{schoenberger2016sfm, schoenberger2016mvs}.
Recent advances have explored augmenting traditional multi-view reconstruction with learned priors, leveraging deep neural networks to improve accuracy and robustness. While these methods often incur higher computational costs and longer training times, they have demonstrated improved performance in challenging reconstruction scenarios. Neural rendering methods, such as Neural Radiance Fields (NeRF) \cite{mildenhall2021nerf} and 3D Gaussian Splatting \cite{kerbl20233d}, have also become dominant. While visually impressive, current approaches are typically slow to train, require many input views, and lack strong geometric priors, limiting their robustness for accurate 3D reconstruction. In our pipeline, we adopt MVSFormer++~\cite{cao2024mvsformer++}, a transformer-based architecture designed to enhance multi-view stereo estimation through learned attention. \\

\noindent
{
\textbf{Pointcloud Shape Completion}: Incomplete or sparse point clouds are a common limitation in 3D reconstruction pipelines, often resulting from occlusions, limited viewpoints, or sensor noise \cite{pauly2005example, yan2022shapeformer}. Shape completion methods aim to infer missing geometry and recover plausible object or scene structures from partial observations \cite{zhuang2024survey}. Traditional approaches rely on geometric priors or optimization-based techniques \cite{pauly2005example, huang2009consolidation}, while recent methods leverage deep learning to learn shape priors from large datasets \cite{yuan2018pcn, wang2020cascaded}. These learned models have demonstrated strong generalization capabilities, particularly in filling in large missing regions with semantically consistent geometry.\\ %TODO: add some actual references here.
}

\noindent
\textbf{Human Foot Reconstruction}: Early attempts in foot modeling relied on Principal Component Analysis (PCA) \cite{amstutz2008pca}, but these models offered limited flexibility with restrictive shape spaces. Later approaches employ active sensor technologies, where structured light or depth cameras are used to generate point clouds \cite{lunscher2017point,yuan20213d,lochner2014development}. However, the point cloud geometries obtained from these sensors are often noisy and incomplete. More recently, Boyne et al. proposed the FIND model \cite{boyne2022find} leveraging a template deformation strategy guided by an implicit neural network to improve reconstruction accuracy. 
Similarly, Osman et al. \cite{osman2022supr} developed SUPR, a PCA-based human foot model designed for seamless integration with the SMPL full-body model \cite{loper2023smpl}, enabling expressive and anatomically consistent reconstructions. 
However, neither FIND nor SUPR are well suited for handling partial or noisy inputs.\\

% {\color{red}

% Our method, draws inspiration from multi-view reconstruction \cite{schoenberger2016sfm,schoenberger2016mvs,kazhdan2006poisson,kazhdan2013screened} and shape completion \cite{hu2019local,shen2012structure}, both of which have proven effective in broader 3D reconstruction tasks. By leveraging these advancements, our approach can be seamlessly integrated into existing works, providing a more robust and generalizable solution for foot reconstruction.}

\begin{figure*}[ht]
    \centering
    \includegraphics[width=0.9\linewidth]{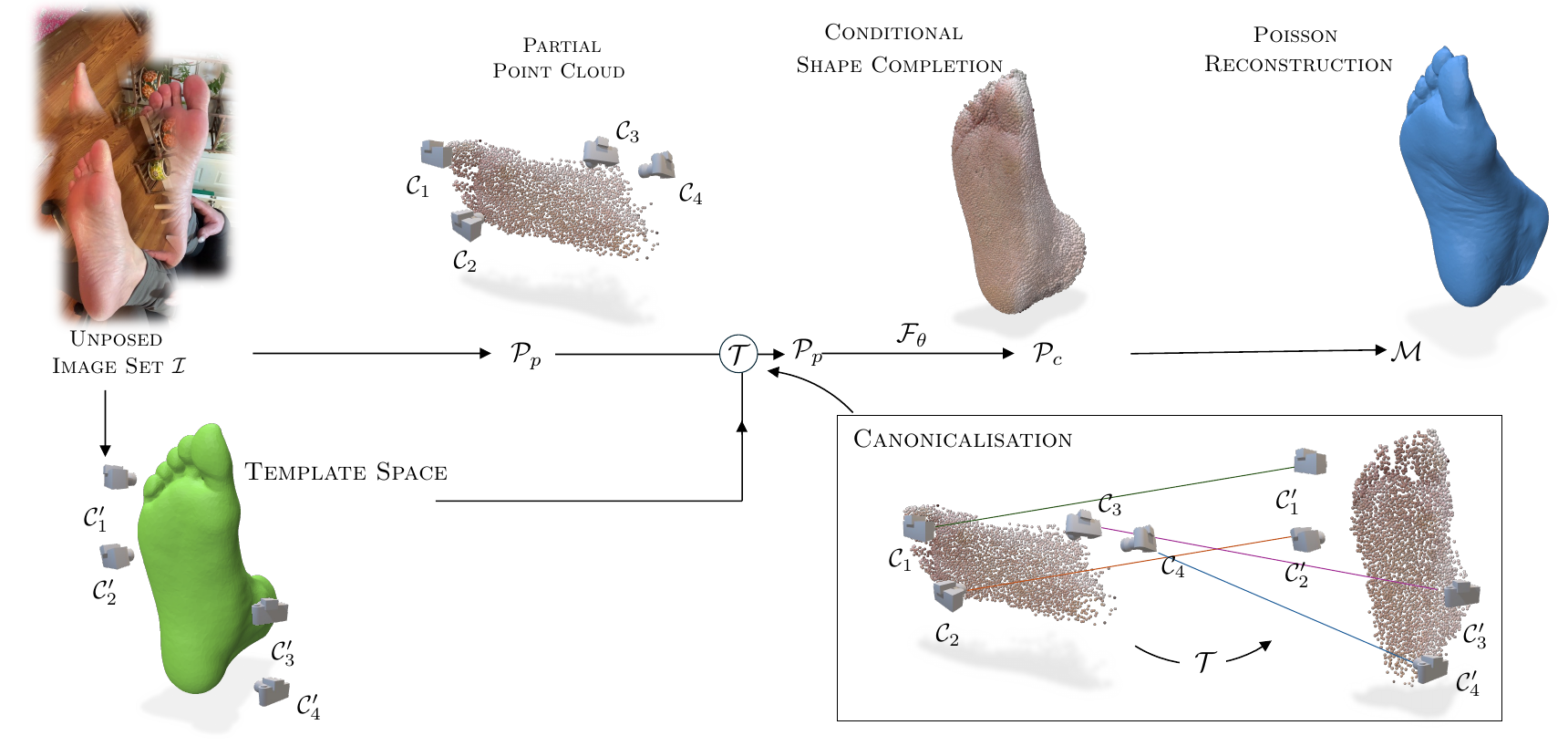}
    \caption{An overview of our reconstruction pipeline: Given an unposed image set $\mathcal{I}$, our method runs two parallel branches: SfM for camera calibration and point cloud generation, and pose estimation. We then canonicalize the point cloud with $\mathcal{T} \in \mathrm{SE}(3)$, apply shape completion $\mathcal{F}_\theta$, and use Poisson reconstruction for the final mesh.}
    %}
    \label{fig:reconstruction_pipeline}
    \vspace{-4mm}
\end{figure*}
% \newpage
\section{Problem Setup}
We consider a set of unposed images of the foot, denoted as \( \mathcal{I} = \{I_1, I_2, \dots, I_N\} \), where each image \( I_i \in \mathbb{R}^{\mathrm{H} \times \mathrm{W} \times \mathrm{C}} \). Our objective is to reconstruct the complete geometry of the foot. To this end, we define a learnable function \( \mathcal{F}_\theta \) that maps the image set \( \mathcal{I} \) to a completed point cloud, such that $
\mathcal{P}_c = \mathcal{F}_\theta(\mathcal{I}).$
To effectively address this, we decompose \( \mathcal{F}_\theta \) into two composite functions $\mathcal{F}_\theta := \mathcal{D}_\theta \circ \mathcal{S},$ 
where \( \mathcal{S}: \mathcal{I} \to \mathbb{R}^{Q \times 3} \) generates a dense, yet potentially incomplete, point cloud of the foot from the unposed images, and \( \mathcal{D}_\theta: \mathbb{R}^{Q \times 3} \to \mathbb{R}^{M \times 3} \) maps this partial point cloud to the completed point cloud target \( \mathcal{P}_c \in \mathbb{R}^{M \times 3} \).\\

The challenge in learning $\mathcal{F}_\theta$ stems from supervision difficulties across inconsistent vector spaces. The geometric transformations between $\mathcal{D}_\theta$ and $\mathcal{S}$ are in general unknown, creating constraints on pose and scale that complicate end-to-end system development. Our key insight is to address this by decomposing the problem into manageable sub-problems and leveraging synthetic training data at each stage. Then by leveraging a robust canonicalization approach we can effectively connect each stage together.  In the following section, we outline our method in more detail.

\section{Method}\label{sec:method}
We tackle complete foot reconstruction with a two-phase approach: first, we use Structure-from-Motion (SfM) and Multi-View Stereo (MVS) to estimate camera pose and generate an initial, though potentially incomplete, point cloud. Then, our shape completion module completes missing geometry to produce a dense, complete representation.\\

A straightforward strategy is to estimate the geometric transformation between the two spaces using Iterative Closest Point (ICP) \cite{besl1992method}. However, this often fails because the point clouds produced by SfM/MVS are typically sparse and incomplete, leading to unreliable alignment. To address this, we introduce a viewpoint prediction (VPP) module that robustly estimates the transformation between the SfM/MVS output and the expected input alignment for shape completion.\\

In the following section, we outline the core components of our reconstruction pipeline, illustrated in Fig.~\ref{fig:reconstruction_pipeline}. The pipeline begins with two branches: {View-Point Prediction} (VPP) and {SfM \& MVS}, discussed in Sec.~\ref{subsec:vpp} and Sec.~\ref{subsec:sfm}, respectively. The VPP module {canonicalizes} the recovered partial point cloud (Sec.~\ref{sec:canon}) before proceeding with {foot completion} and reconstruction (Sec.~\ref{sec:plccomplete}).

\subsection{View-point Prediction}\label{subsec:vpp}
The first branch of our architecture, the VPP module, estimates both a bounding box of the foot and the pose relative to a predefined template mesh. Given an unposed image set $\mathcal{I}$ and a reference mesh $\mathcal{M}_{\mathrm{ref}}$, we train a neural network to regress the approximate six degrees of freedom (6-DoF) of the camera pose relative to $\mathcal{M}_{\mathrm{ref}}$. Our method builds on YOLO6D \cite{maji2024yolo}, adopting a similar training strategy and leveraging synthetic data; implementation details are in Sec.~\ref{implementation}. We represent the VPP module as $\mathcal{V}_{\phi}$ and define its output for a given image $\mathcal{I}_i \in \mathcal{I}$ as: $(\hat{\mathcal{C}}_i, B_i) = \mathcal{V}_{\phi}(\mathcal{I}_i),$ where $\hat{\mathcal{C}}_i$ denotes the estimated camera parameters, and $B_i$ represents the bounding box of the foot in the image.

\subsection{SfM \& MVS}\label{subsec:sfm}
We use a standard structure-from-motion (SfM) pipeline to estimate 3D structure by matching keypoints across views and jointly refining camera poses and a sparse point cloud via bundle adjustment.
% We employ a standard structure-from-motion (SfM) pipeline, which estimates 3D structure by matching keypoints across views and refining camera poses and a sparse point cloud via bundle adjustment, jointly across all images. 
Specifically, we utilize GLOMAP \cite{pan2024glomap}, which from our experiments, observed to give significantly more efficient and scalable global reconstruction compared to COLMAP \cite{schoenberger2016sfm,schoenberger2016mvs}. For the image-set $\mathcal{I}$, we model the SfM process as:
$$
\underbrace{\{\mathcal{C}_1, \cdots, \mathcal{C}_N\}}_{\boldsymbol{\mathcal{C}}} = \mathrm{SfM}(\mathcal{I}),
$$
where each $\mathcal{C}_i \in \boldsymbol{\mathcal{C}}$ represents the estimated camera parameters of image $\mathcal{I}_i$. Using the bounding box $B_i$ from the VPP module, we generate segmentation masks via the Segment Anything Model 2 (SAM2) \cite{ravi2024sam2} in a zero-shot setting. %(see Fig.~\ref{fig:segmentation}). 
Denoting the set of all bounding box centers as $\mathcal{B}$, we model the segmentation process as:  
$$
\underbrace{\{\hat{\mathcal{I}_1}, \cdots, \hat{\mathcal{I}_N}\}}_{\boldsymbol{\hat{\mathcal{I}}}} = \mathrm{SAM}(\mathcal{I}, \mathcal{B
}),
$$ where $\hat{\mathcal{I}_1}$ is the $i$-th masked image. 
To reconstruct a dense point cloud, we employ a multi-view stereo (MVS) approach~\cite{Hartley2004}, which estimates depth by matching pixel correspondences across multiple views and refining depth maps; in this work we leverage the state-of-the-art MVSFormer++~\cite{cao2024mvsformer++} to recover a high-quality point-cloud. Using the camera parameters from GLOMAP and segmentation masks from SAM2, we reconstruct the visible foot geometry pointcloud as:
$$
\mathcal{P}_p = \mathrm{MVS}(\mathcal{I}, {\mathcal{C}}, {\hat{\mathcal{I}}}).
$$
% To reconstruct a dense point cloud, we employ a multi-view stereo (MVS) approach \cite{Hartley2004}, leveraging the state-of-the-art MVSFormer++ \cite{cao2024mvsformer++} for high-quality point-cloud generation. 
% \begin{figure}[ht]
%   \centering
%   \begin{subfigure}[b]{0.30\linewidth}
%     \centering
%     \includegraphics[width=\linewidth]{figures/segmentation/input.jpg}
%     \caption{}
%     \label{fig:seg_input}
%   \end{subfigure}
%   \hfill
%   \begin{subfigure}[b]{0.30\linewidth}
%     \centering
%     \includegraphics[width=\linewidth]{figures/segmentation/binary.jpg}
%     \caption{}
%     \label{fig:seg_binary}
%   \end{subfigure}
%     \hfill
%   \begin{subfigure}[b]{0.30\linewidth}
%     \centering
%     \includegraphics[width=\linewidth]{figures/segmentation/seg.jpg}
%     \caption{}
%     \label{fig:seg_masked}
%   \end{subfigure}
%   \caption{(a) Input image to the segmentation mask. (b) Binary mask produced by SAM2 (c) Image masked to region of interest.}
%   \label{fig:segmentation}
% \end{figure}

\subsection{Point Cloud Canonicalization}\label{sec:canon}

Partial point-clouds recovered from image sets have arbitrary poses and scales, complicating their use in downstream shape completion. To address this, we transform the point clouds into a known canonical frame using the camera parameters $\hat{C}_i$ estimated by the VPP module and the depth maps estimated in the MVS process $\mathcal{D}_i$. We present our canonicalization approach in more detail in Algorithm. 1.%\ref{ref:alg1}.

\subsection{Point Cloud Completion}\label{sec:plccomplete}

The second stage of our robust reconstruction pipeline focuses on completing the foot geometry using the learned function $\mathcal{D}_\theta ({\mathcal{P}})$. For this, we propose an attention-based point cloud completion framework that operates on partially reconstructed foot geometries from the SfM/MVS stage.\\

Building on recent attention-based approaches to point cloud modeling \cite{wang2024pointattn}, we formulate completion as an auto-encoding problem, where the model predicts a global latent representation to guide the reconstruction. Our attention mechanism aggregates information across the entire point cloud, capturing both local details and global structural patterns without relying on predefined neighborhood structures. We adopt a coarse-to-fine reconstruction strategy with a scaffold-based skip connection that directly integrates a subset of the input point cloud into the reconstruction process. This scaffold helps maintain fidelity to the observed geometry while enabling the model to infer missing regions effectively and reduce scan noise.\\

In our standard pipeline, we then employ the screened Poisson surface reconstruction (SPSR) algorithm \cite{kazhdan2013screened} to generate a mesh, with normals estimated via PCA.\\

% In the next section we provide an overview of the implementation of our method. 

\begin{algorithm}
\caption{Canonicalization}
\begin{algorithmic}[1]
\REQUIRE Reference mesh: $\mathcal{M}_{\text{ref}}$,\\
~~~~~~~~~~Camera parameters: $C_i$ [SfM], $\hat{C}_i$ [VPP],\\
~~~~~~~~~~Depth maps: $D_i$, Point-cloud: $\mathcal{P}_p$.
\STATE Select $k$ points $p_1, p_2, \ldots, p_k$ from $\mathcal{M}_{\text{ref}}$
\FOR{$i \leftarrow 1$ to $N$}
    \FOR{$j \leftarrow 1$ to $k$}
        \STATE $q_{i,j} \leftarrow \text{Project}(p_j, \hat{C}_i)$
        \STATE $d_{i,j} \leftarrow D_i(q_{i,j})$
        \STATE $p'_{i,j} \leftarrow \text{BackProject}(q_{i,j}, d_{i,j}, C_i)$
    \ENDFOR
\ENDFOR
\STATE Compute Centroids $c_j$ for each $j = 1 \ldots k$
\STATE Procrustes: $\{R, t, s\} \leftarrow \text{Procrustes}(\{p_j\}, \{c_j\})$
\STATE $\mathcal{P}'_p \leftarrow s(R \cdot \mathcal{P}_p + t)$
\STATE ICP refinement: $\{R', t'\} \leftarrow \text{ICP}(\mathcal{P}'_p, \mathcal{M}_{\text{ref}})$
\STATE $\mathcal{P}_{\text{aligned}} \leftarrow R' \cdot \mathcal{P}'_p + t'$
\RETURN $\mathcal{P}_{\text{aligned}}$
\end{algorithmic}
\end{algorithm}

\section{Implementation}\label{implementation}

In this section we provide details of the implementation and training procedure used to construct our reconstruction pipeline.\\

\begin{figure}
    \centering
    \includegraphics[width=0.85\linewidth]{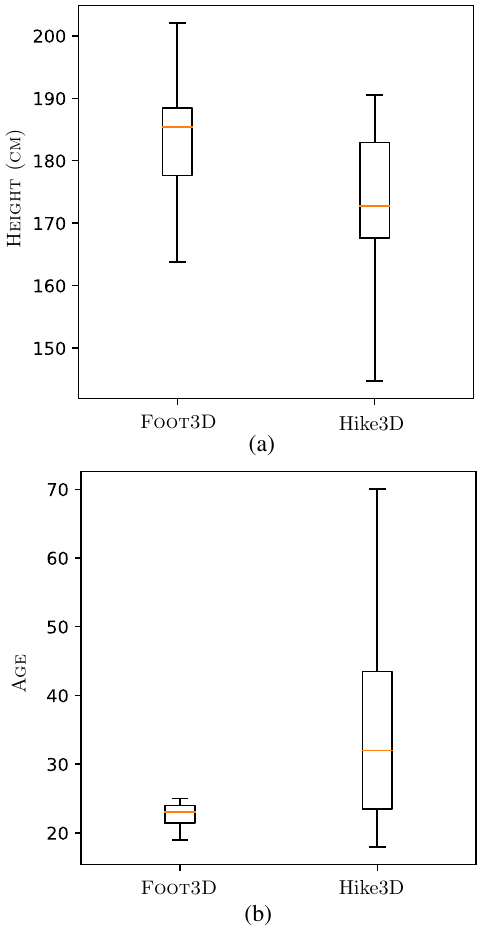}
    \caption{Boxplots summarizing the distribution of heights and ages in our dataset, \texttt{Hike3D}, reveal a broader diversity in heights and significantly greater variation in ages compared to prior datasets. This age variation is particularly important, as feet undergo changes and deformations over time due to factors like footwear and activity levels.}
    \label{fig:boxplots}
\end{figure}

\noindent
\textbf{Datasets}: High-fidelity foot geometry datasets are scarce.  Foot3D \cite{boyne2022find} is a valuable resource, but its narrow age and height range prompted us to develop \texttt{Hike3D}, a more diverse dataset for orthotics research. The \texttt{Hike3D} dataset comprises 15 participants recruited in the United States from a variety of occupational backgrounds. Data acquisition was performed using an EinStar structured-light scanner in a controlled indoor setting. Participants were seated with one foot extended, barefoot, in a neutral posture. Each foot was scanned individually, beginning from the plantar surface and sweeping around to the dorsal surface, capturing full geometry from bottom to top. Scans were cleaned, aligned to a common reference frame, and manually inspected for quality control, with their geometric accuracy evaluated in an offline assessment. To broaden demographic coverage and strengthen design robustness, we integrate \texttt{Hike3D} with Foot3D. We provide an overview of  the distribution  of \texttt{Hike3D} compared with \cite{boyne2022find} in Fig.~\ref{fig:boxplots}. We will release the dataset as part of this work.\\

% \subsection{Training Procedure}

\noindent
\textbf{VPP module}: We train the VPP module using synthetically generated images of meshes from our training set. To ensure diversity, we utilize 740 HDR backgrounds, creating various background combinations for our 50k synthetic images. We use Blender to simulate lighting variations and skin tone modulation, enhancing the realism of the synthetic data. We then fine-tune the model using 5k real images. Throughout the process, we apply the same augmentations and loss functions as in the original work \cite{maji2024yolo}.\\

\noindent
\textbf{Foot Completion Module}: We train the foot completion module using a simulated scanning setup to generate paired partial and complete geometries. To improve robustness against noise and SE(3) perturbations, we apply data augmentations during training. Our dataset combines \texttt{Hike3D} and Foot3D, splitting the dataset with 80\% for training and 20\% for testing. Each mesh was augmented with 10 spatial transformations (shifts, scaling, rotations), followed by five virtual scans per transformation, yielding 2000 training and 250 testing pairs. Supervision is applied by minimizing the Chamfer distance between predicted and ground-truth point clouds at intermediate steps of the network.

\section{Experiments}
To evaluate robustness, we conduct two experiments: one using paired videos and high-quality 3D scans for quantitative error analysis, and another using video-only data, where clinicians score reconstructions based on visual assessment.
\begin{figure*}[ht]
  \centering
  \includegraphics[width=\textwidth]{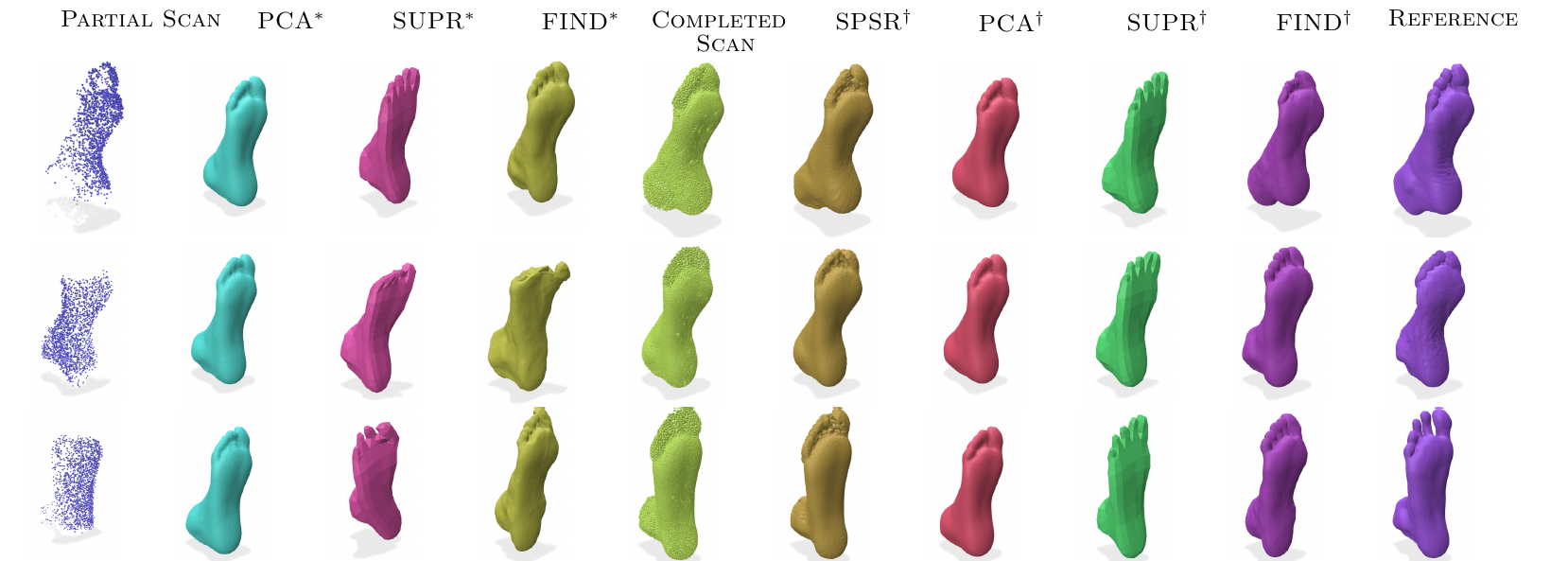}
  \caption{Figure shows partial scan reconstruction results. Methods marked $*$ are optimized on the input scan, while $\dagger$ denotes optimization on our completed point cloud, which recovers geometry much closer to the reference scans.
}
    % \vspace{-4mm}
  \label{fig:7}
\end{figure*}
\subsection{Experimental Setup}

\begin{figure}[ht]
\centering
\begin{minipage}[t]{0.5\textwidth}
  \vspace{0pt}%
  % Your table
  \centering
  \begin{tabular*}{\linewidth}{@{\extracolsep{\fill}}lcc}
    \toprule
    \textsc{Method} & CD ($\downarrow$)~($10^{-2}$) & HD ($\downarrow$)~($10^{-2}$)\\
    \midrule
    PCA & $4.46 \pm 1.24$ & $12.28 \pm 3.26$ \\
    SUPR & $12.74 \pm 3.78$ & $34.61 \pm 7.91$ \\
    FIND & $15.95 \pm 6.11$ & $37.19 \pm 14.36$ \\
    \midrule
    Ours & $2.29 \pm 0.56$ & $9.51 \pm 3.76$ \\
    \midrule
    SPSR + Ours& $2.81 \pm 0.77$& $10.20 \pm 3.13$ \\
    PCA  ~+ Ours & $3.93 \pm 1.08$ & $11.37 \pm 3.02$ \\
    SUPR + Ours & $7.08 \pm 1.79$ & $27.95 \pm 5.43$ \\
    FIND + Ours & $3.46 \pm 1.26$ & $10.05 \pm 3.50$ \\
    \bottomrule
  \end{tabular*}
  \captionof{table}{Average Chamfer Distance (CD) and Hausdorff Distance (HD) results. The top section shows results from direct fitting to partial inputs, middle is results for our generated point cloud, and bottom is results from fitting on our generated point clouds. Reported values include $\pm$1 standard deviation over the test dataset.} %Average Chamfer Distance (CD) and Hausdorff Distance (HD) results. The top section shows models directly fitted to partial inputs. The middle section presents our reconstructed output point clouds. The bottom section reports results from fitting each model to our output point clouds. Reported values include ±1 standard deviation over the test dataset.}
  \label{tab:fitting_methods}

  \vspace{6pt}%
\end{minipage}
\hfill
\begin{minipage}[t]{0.44\textwidth}
  \vspace{0pt}%
  % Your figure
  \centering
  \includegraphics[width=0.90\linewidth]{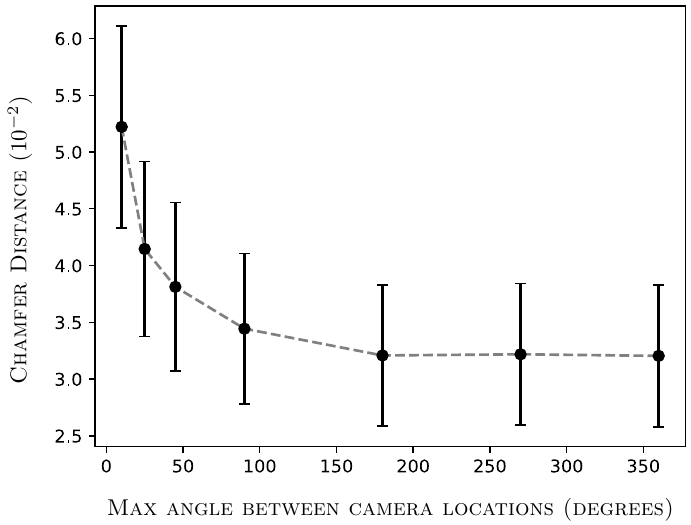}
  \captionof{figure}{Chamfer Distance between predicted foot mesh and ground truth vs. camera scanning angle.}
  \label{fig:chamfer_dist_vs_angle}
\end{minipage}
\end{figure}

\noindent
\textbf{Completion Module Evaluation}: To assess the efficacy of our foot completion module, we perform a quantitative comparison against three established foot modeling techniques. Each method is tasked with fitting a model to both the original incomplete scans and our completed point clouds. The baseline methods are: (1) a PCA-based model \cite{amstutz2008pca} with vertex correspondences obtained using functional maps \cite{ovsjanikov2012functional,melzi2019zoomout}; (2) the parametric SUPR foot model \cite{osman2022supr}; and (3) FIND \cite{boyne2022find}, which utilizes a large latent space for detailed shape and pose control. For all methods, we optimize shape, pose, and transformation parameters via gradient descent with the Adam optimizer \cite{kingma2014adam}, minimizing the Chamfer distance. Final reconstruction accuracy is quantified using both Chamfer and Hausdorff distances.\\

% \noindent
% \textbf{Completion Module}: To evaluate our foot completion module, we fit three established foot models to incomplete and completed scans. The first, a PCA-based method~\cite{amstutz2008pca}, uses functional maps~\cite{ovsjanikov2012functional,melzi2019zoomout} for vertex correspondences and fits a PCA model to mesh displacement vectors. The second leverages the SUPR foot model~\cite{osman2022supr}, while the third, FIND~\cite{boyne2022find}, offers a large latent space for shape and pose control, with added transformation parameters for better alignment. For all methods, we optimize shape, pose, and transformation parameters via gradient descent, minimizing Chamfer distance with the Adam optimizer~\cite{kingma2014adam}. Final accuracy is assessed using Chamfer and Hausdorff distances.\\

\noindent
\textbf{End-to-End Pipeline Evaluation}: We evaluate the performance of our entire reconstruction pipeline using video footage captured in uncontrolled, real-world settings. Our method is benchmarked against two prominent 3D reconstruction pipelines: (1) COLMAP \cite{schoenberger2016sfm}, a standard Structure-from-Motion (SfM) and Multi-View Stereo (MVS) pipeline that reconstructs dense geometry, and (2) Gaussian Opacity Fields (GOF) \cite{yu2024gaussian}, a state-of-the-art method for adaptive surface reconstruction from images. Our dataset consists of 30 videos from smartphone videos, featuring diverse subjects, lighting, and backgrounds. For qualitative assessment, three expert clinicians in foot anatomy and orthotics reviewed randomized renderings from all three methods. They rated each reconstruction on a 5-point Likert scale based on three criteria: (1) anatomical accuracy, (2) completeness of the foot structure, and (3) suitability for orthotic design.
\subsection{Experimental Results}

\textbf{Completion Module Performance}: As detailed in Table \ref{tab:fitting_methods}, our point cloud completion method significantly outperforms the template-based fitting approaches when applied to incomplete data, achieving lower Chamfer and Hausdorff distances. Furthermore, when the outputs are meshed using Screened Poisson Surface Reconstruction (SPSR), the surfaces generated from our completed point clouds exhibit the lowest Chamfer distance, affirming the benefit of our approach for foot mesh recovery. The qualitative results in Figure \ref{fig:7} visually corroborate these quantitative improvements, showing more plausible and complete foot geometry.\\

% \textbf{Point Completion}: Table \ref{tab:fitting_methods} demonstrates that our point cloud-based method consistently outperforms template-based approaches on incomplete point clouds, yielding lower Chamfer and Hausdorff distances. Furthermore, meshes reconstructed through our pipeline, when meshed using SPSR, achieve the lowest Chamfer distances among all meshed, further validating our design choice. 
% Qualitative results in Figure \ref{fig:7} further illustrate these improvements.\\

% \vspace{-4mm}

\noindent
% \textbf{View-Completeness:} We evaluate our shape completion module under varying levels of partialness using a simulated scanning setup. By limiting the maximum angle between the camera-to-foot vector, we control foot coverage—smaller angles mean more missing data. As shown in Fig.~\ref{fig:chamfer_dist_vs_angle}, increasing this angle lowers the Chamfer Distance, improving reconstruction accuracy. Notably, the error plateaus around 90$^{\circ}$, a feasible range for a person performing self-scanning, indicating that this range balances practicality and accuracy, further validating our design choice for a robust completion module.\\
\textbf{Robustness to Partial Views}: We analyzed our shape completion module's performance under varying degrees of data incompleteness. In a simulated scanning environment, we controlled the partialness of the input scan by restricting the maximum viewing angle between the virtual camera and the foot surface. As shown in Figure \ref{fig:chamfer_dist_vs_angle}, the reconstruction error (Chamfer Distance) systematically decreases as the viewing angle—and thus the surface coverage—increases. Notably, the error rate plateaus around 90$^{\circ}$, a viewing range that is practically achievable in self-scanning scenarios. This demonstrates that our module delivers robust and accurate completions.\\
\begin{figure}[h!]
    \centering
    \includegraphics[width= 0.72\linewidth]{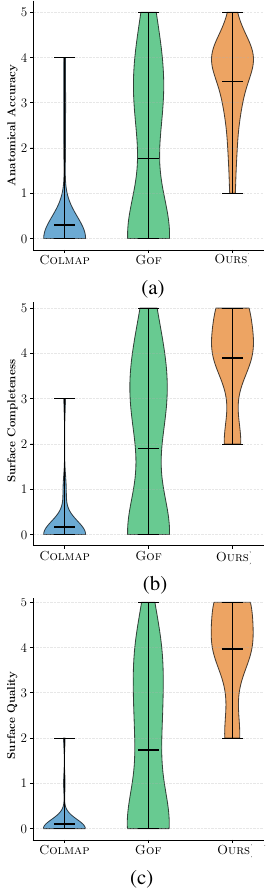}
    \caption{Plots of the distribution of aggregated clinical scores for each methods assessing (a) anatomical accuracy (b) completeness (c) surface quality.}
    \label{fig:6}
\end{figure}
% \begin{figure*}[ht]
%   \centering
%   \includegraphics[width=\textwidth]{figures/clinical_final.pdf}
%   \caption{Plots of the distribution of aggregated clinical scores for each methods assessing (a) anatomical accuracy (b) completeness (c) surface quality.}
%     % \vspace{-4mm}
%   \label{fig:6}
% \end{figure*}

% \textbf{End-to-End Pipeline}: We present our results in Fig.~\ref{fig:6} (a)–(c). Our method consistently outperforms COLMAP and GOF across all metrics. COLMAP shows the worst anatomical accuracy with major deviations, while GOF is inconsistent. Our method achieves the highest, most consistent ratings in fidelity, completeness, and surface quality. These results confirm its suitability for clinical applications like foot orthotic design and precision insole manufacturing.
% \newpage
\noindent
\textbf{End-to-End Clinical Assessment}: The results of our end-to-end evaluation are summarized in Figure \ref{fig:6} (a)–(c). The clinical assessment reveals that our method consistently outperforms both baselines across all criteria. Reconstructions from COLMAP were frequently cited for poor anatomical accuracy and major geometric deviations. While GOF performed better, its results were rated as inconsistent. In contrast, our method achieved the highest and most stable scores for anatomical fidelity, completeness, and suitability for orthotics. These findings underscore our system's potential for reliable use in clinical applications, such as the design and manufacturing of custom foot orthotics.\\

\section{Discussion}
Our results demonstrate that our end-to-end pipeline significantly improves foot geometry reconstruction, achieving lower Chamfer and Hausdorff distances while maintaining consistency with input data. The foot completion module, leveraging learned priors, successfully reconstructs plausible geometries from sparse data, addressing the limitations of template-based methods, which showed constrained shape variability in our evaluations. Our approach enables robust reconstruction across diverse and incomplete inputs, as reflected in its consistently high surface completeness scores. Furthermore, by integrating completion and canonicalization, our method effectively mitigates occlusions and partial view challenges, leading to more accurate and reliable reconstructions, as evidenced by its superior anatomical fidelity and quality ratings. While overall performance is strong, fine structures such as toes can be more challenging to reconstruct accurately, particularly in sequences where the static scene assumption is mildly violated due to subtle motion. Additionally, the current model lacks uncertainty quantification for highly out-of-distribution geometries, which we plan to explore in future work.\\

% While we have shown our model performs well across a diverse range of test cases, it does lack explicit uncertainty quantification for extreme out-of-distribution foot geometry; we will seek to address this in future work.

% % 
% \begin{figure*}
%     \centering
%     \includegraphics[width=\linewidth]{figures/simulated_scannign.png}
%     \caption{Captures of the simulated scanning setup used to train the shape completion module. For each mesh we construct a hemisphere of cameras and select a subset of the of them to project points onto the foot mesh. The points are then down-sampled to the correct cardinality and further augmentations like Gaussian additive noise can be applied.}
%     \label{fig:simulated_scanning}
% \end{figure*}

\section{Conclusion}

We introduced a novel end-to-end pipeline for reconstructing foot geometry from self-scanned mobile videos, addressing key limitations of existing methods. Our proposed method provides robust foot reconstruction, even from partial observation. Extensive evaluation demonstrated that our method outperforms baseline approaches, achieving lower Chamfer and Hausdorff distances while preserving consistency with input geometry. These findings underscore the effectiveness and robustness of our approach, particularly for self-scanning applications, paving the way for improved foot reconstruction in real-world settings.  

\section*{Acknowledgments}
This work was supported by the Engineering and Physical Sciences Research Council [EP/S023917/1].
\clearpage
\newpage
{
    \small
    \bibliographystyle{ieeenat_fullname}
    \bibliography{main}
}

\end{document}